\documentclass[journal,onecolumn]{IEEEtran}
\usepackage{amsmath,amsfonts}
\usepackage{algorithmic}
\usepackage{algorithm}
\usepackage{array}
\usepackage[caption=false,font=normalsize,labelfont=sf,textfont=sf]{subfig}
\usepackage{textcomp}
\usepackage{stfloats}
\usepackage{url}
\usepackage{verbatim}
\usepackage{graphicx}
\usepackage{cite}
\usepackage{gensymb}
\usepackage{color,soul}

\newcommand{\norm}[1]{\left\lVert#1\right\rVert}

\begin{document}

\title{Cluster \& Disperse: a general air conflict resolution heuristic using unsupervised learning}

\author{Mirmojtaba~Gharibi$^a$ 
        and~John-Paul~Clarke$^b$
        \\$^{a,b}$Department of Aerospace Engineering and Engineering Mechanics, 
        \\$^{a,b}$The University of Texas at Austin. 
        \\$^a$mgharibi@uwaterloo.ca
}
\markboth{}%
{Shell \MakeLowercase{\textit{et al.}}: A Sample Article Using IEEEtran.cls for IEEE Journals}

\IEEEpubid{}

\maketitle

\textbf{Abstract}:
We provide a general and malleable heuristic for the air conflict resolution problem. This heuristic is based on a new neighborhood structure for searching the solution space of trajectories and flight-levels. Using unsupervised learning, the core idea of our heuristic is to cluster the conflict points and disperse them in various flight levels. Our first algorithm is called Cluster \& Disperse and in each iteration it assigns the most problematic flights in each cluster to another flight-level. In effect, we shuffle them between the flight-levels until we achieve a well-balanced configuration. The Cluster \& Disperse algorithm then uses any horizontal plane conflict resolution algorithm as a subroutine to solve these well-balanced instances. Nevertheless, we develop a novel algorithm for the horizontal plane based on a similar idea. That is we cluster and disperse the conflict points spatially in the same flight level using the gradient descent and a social force. We use a novel maneuver making flights travel on an arc instead of a straight path which is based on the aviation routine of the Radius to Fix legs. Our algorithms can handle a high density of flights within a reasonable computation time. We put their performance in context with some notable algorithms from the literature. Being a general framework, a particular strength of the Cluster \& Disperse is its malleability in allowing various constraints regarding the aircraft or the environment to be integrated with ease. This is in contrast to the models for instance based on mixed integer programming.


\section*{\begin{flushleft}\textbf{Introduction}\end{flushleft}}

Many of the algorithmic advances in combinatorial optimization problems have been the result of new innovative structures for search neighborhood. Air conflict resolution, being a highly combinatorial problem, should also benefit from these novel structures. 
However, there is no widely accepted  search neighborhood structure for conflict resolution models such that the research community can easily integrate existing algorithms with such a framework.

With the growth of air traffic and sector capacity overloads, there has been a plethora of models for solving aircraft conflict resolution problem (CRP) in a horizontal plane. However, to be fully operational, these models need to be integrated in a way that utilize the vertical flight-levels. Cluster \& Disperse is a fast and general heuristic based on unsupervised learning that extends the planar conflict resolution models to 3D.





There has been much prior work on planar conflict resolution. One problem in those works, as will be discussed is that the instances of the flights given to the algorithm are overly difficult instances. The main approach used for reducing the complexity of these instances are to reject (i.e., remove) some of the flights that do not fit well with a partial conflict resolution solution. Even the instances that contain flights flying in the same general direction are still difficult due to the number of crossing trajectories.


Another problem lies with the conflict resolution models themselves, especially those based on mixed integer programming (MIP). Many of these models have overly rigid or bloated number of constraints such as assuming identical speed, allowing only one change of heading, etc. This makes them less adaptable to the practical situations.


However, we believe we can address both of these problems as will follow. Firstly, if we take a more global view, we observe that for every flight level, the flight instances given to the planar conflict resolution solver is difficult. It is possible to assign flights to different flight levels such that the complexity of instances given to almost all planar CRP solvers is reduced. This way, they will all achieve a better result. To use an analogy for the problem and our solution, this is akin to a game of Tetris (analogous to conflict resolution on each flight level) where the random pieces don't fit well together. However, if we are allowed to choose to which game the new piece will be added, we can add it to the game where pieces fit better. That way, all of the games will be easier. Secondly, we use a generic approach that can readily admit new constraints. For example, each aircraft can have a different model, or might only be assigned to certain flight levels.


To comment on our approach in a more technical sense, we reassign the flights on each previously tentatively assigned flight-level according to the conflict clusters that are formed. We then attempt to disperse each formed cluster into other flight-levels and create more balanced input instances for each planar CRP solver. This cluster-based neighborhood structure forms the basis of our approach.


\subsection*{\textbf{1. Contributions}}
Our contribution can be summarized as follows.
We develop a novel neighborhood structure for the local search for conflict-free trajectories based on clusters of conflict formed at each flight-level. Based on this neighborhood structure, we develop a new general heuristic for assigning conflict free flight-level and trajectories. Our heuristic can use any planar conflict resolution algorithm as a subroutine, as long as the model can provide partial solutions in the case a completely conflict-free trajectory does not exist.

While we can use any planar CRP solver as a subroutine, we also developed a planar conflict resolution method based on clustering. In particular, the novelties of this planar CRP solver are two folds. Firstly, we perform the clustering in a novel way, using a novel definition for data points. We formulate traffic clusters in a new way where our data points that form the clusters are the minimum distance points of trajectories. Secondly, to the best of our knowledge, the existing clustering methods in the context of conflict resolution are all rule-based. This is the first time that clustering is performed using the unsupervised learning methods in this context. 

Furthermore, unlike the existing literature, we use RF-legs between the entry and exit points of the sector. By changing the radius of the turn (or equivalently the arc angle), we bend the flight trajectories from the straight path and avoid conflicts. In effect, we disperse the clusters locally by making the cluster expand as a result of the social force from the points in the cluster. The RF-legs are simple routine procedures that can be performed by the pilot by setting the bank angle. To the best of our knowledge, this is the first time RF-leg maneuvers are being used as a systematic way for ensuring separation.

In terms of performance, each iteration of the Cluster \& Disperse takes a negligible amount of time compared to the run-time of the planar CRP solvers (less than 10s for the initialization, and less than 0.01s per iteration). Therefore, the given performance makes this framework suitable for most real-time applications.

A great strength of our framework is its malleability and the ease with which new constraints can be integrated. 

Finally, the foremost concern for the traffic controllers is that of maintaining the separation and ensuring safety. As such, complexities, like the flight efficiency at different levels and the cost of maneuvers are addressed at a different organizational level and hence are not the concern of our paper. Since our heuristic is a very malleable one, other limitations such as climbing limitations or the maximum deviation from the straight path for a particular aircraft can be added as a set of constraint in the planar CRP solver (based on RF-legs) and the flight level assignment. For example, if for an aircraft we must choose a particular flight-level, we assign the flight-level accordingly for that aircraft and choose other aircraft for being assigned to other flight levels in order to disperse conflicts as will be explained in more details in the upcoming sections. 

\subsection*{\textbf{2. Related work}}

The work in \cite{lehouillier2017two} models the conflict resolution problem as a large neighborhood search where a configuration is related to another if the assigned flight-levels of aircraft are close. In the next step of the process, the process optimizes the horizontal plane for each flight-level. The possible maneuvers for aircraft are a change of heading and speed for the horizontal plane optimization and change of flight-level for the large neighborhood search. The horizontal plane conflict resolution problem is formulated as a graph where the vertices are the maneuvers of each aircraft and those maneuvers that are compatible are connected with an edge. The cost of each edge is the combined cost of associated maneuvers. The goal is to find a minimum cost and maximum cardinality clique (to include the largest number of aircraft). This problem is translated to a sequential mixed integer programming model. It is then solved as a sub-problem within the large neighborhood search framework. 

Some works in the literature use the idea of clustering aircraft in conflict. 
The authors in \cite{granger2001optimal} form a graph of aircraft in conflict within the considered time window where nodes represent the aircraft and the edges represent a pair-wise conflict. This graph represents a conflict cluster. The conflict resolution process will be applied to each separate cluster. If this causes new conflicts, new clusters are formed by merging the clusters and running the conflict resolution process again on the newly formed cluster. This process is repeated until all clusters are conflict-free. The authors in \cite{yang2017two} use a similar graph notion of conflicts but in the context of unmanned aerial vehicles (UAVs) and use heading or speed changes for conflict resolution.

A drawback of this common way of defining clusters is that an aircraft that would have conflicts with two aircraft that are far apart would force them to be in the same cluster unless the time window of consideration is very small. On the other hand, a small time window would cause the clusters to be very small and hence ineffective. Therefore, we use a different definition for our clusters based on the minimum distance points between aircraft. 

The authors in \cite{cafieri2014aircraft} use the mentioned conflict graph to form clusters of at most four aircraft. The authors solve each cluster using mixed integer programming by assigning speed to each aircraft. The whole process is repeated in iterations until all conflicts are resolved and no new conflicts are created in the process. In each iteration, the starting point for the speeds will be the previously allocated speeds to preserve the progress made toward the solution. 

Using a similar definition for a cluster, in \cite{chiang1997geometric}, a box is formed to contain the cluster (i.e. the graph) in addition to a small margin allocated for the so-called buffer aircraft that are just outside the box. These aircraft are treated as obstacles for the conflict resolution algorithm. The way-points for trajectories are then created such that the conflicts are resolved. Similarly, the authors in \cite{krozel2001system} use the idea of bounding boxes and buffer aircraft and a look-ahead time of 8 minutes. The conflicts are resolved in the horizontal plane using only heading changes. To resolve the conflicts in each box, a random ordering of the aircraft is chosen, and the heading for each aircraft is fixed one by one. Each aircraft with a set heading will then be treated as a constraint for the next aircraft. The process is started again with a different ordering if the particular ordering fails to resolve the conflicts. The authors in \cite{fan2016novel} also use a similar method of clustering for UAVs and solve each cluster in parallel for conflict resolution via altitude change maneuvers.
In \cite{dougui2013light}, the authors use a rectangular cuboid around the aircraft in conflict and any trajectory that falls inside this box will be treated as either a conflict or a constraint in case of a non-conflict. Conflict resolution for clusters is then computed in parallel.

The authors in \cite{radanovic2018surrounding}, form clusters of aircraft by matching the closest points of approaches that fall within a bounding box for the cluster within a look ahead time window. A heading change maneuver or an altitude change of $1000ft$ is then considered for conflict resolution.

Various works involve the use of mathematical programming for the conflict resolution of aircraft. A recent survey on these approaches can be found in \cite{pelegrin2022aircraft}. The authors in \cite{pallottino2002conflict} devise a MIP model for the conflict resolution problem in a horizontal plane. They consider the cases of only allowing heading change or speed change separately as possible maneuvers. All maneuvers by the aircraft will be performed at the time $t=0$ only. The authors in \cite{vela2010near} extend this work by allowing simultaneous heading and speed change maneuvers as well as considering objective functions such as the fuel cost associated with the deviation from the flight path. 
In \cite{vela2009mixed}, the authors use a mixed integer programming formulation for time metering the aircraft at intersections flying over predetermined paths. As part of  the maneuver that is needed for achieving the minimum separation, the speed of aircraft or their flight-levels is adjusted. However, the devised model only enforces the separation at the way-points. The authors in \cite{hassan2021mixed} follow a similar approach while taking into account the adjacent sectors.
In \cite{lehouillier2017solving}, the authors develop a mixed integer programming model for conflict resolution where the speed and heading changes are used as the maneuvers of choice for separation. However, these changes are not applied instantaneously in the model as this common simplifying assumption is shown to cause large errors in the estimation of the separation distance. Various uncertainties are considered in the model such as wind prediction errors, aircraft speed measures, and delays in performing the maneuvers. 
The authors of \cite{wang2020cooperation} use a hybrid approach of using a Memetic algorithm and a MIP-based algorithm. Their algorithm allows heading, speed, and flight-level changes for conflict resolution. The authors argue that their hybrid approach outperforms the individual algorithms in solving the conflict resolution problem. 

There are various papers on conflict resolution that take a machine-learning approach. The authors in \cite{pham2019machine} take a reinforcement learning approach based on Deep Q-learning and Deep Deterministic Policy Gradient algorithms to solve the conflict resolution problem. Any maneuver that achieves separation is rewarded and otherwise punished. Each maneuver specifies the time of a heading change and the duration for which the new heading will be kept. The authors in \cite{lai2021multi} formulate the conflict resolution problem as Multi-agent Markov Decision Processes (MMDPs) and use multi-agent reinforcement learning to learn the needed maneuvers which consist of finding intermediate way-points.

There are a variety of other approaches to conflict resolution as well, such as the use of Ant Colony Optimization, Satisficing Approach, etc. (\cite{durand2009ant,archibald2008satisficing, rey2016subliminal, tang2021automated, hwang2007protocol, menon1999optimal}. For a more in-depth and comprehensive review of the literature on the conflict resolution problem, interested readers are referred to \cite{pelegrin2022aircraft}, \cite{ ribeiro2020review, tang2019conflict, martin2010collision, kuchar2000review}.

\section*{\begin{flushleft}\textbf{Problem definition}\end{flushleft}}
\textbf{Conflict Resolution Problem (CRP)}: In this problem, we are given the followings: a sector, designated by horizontal planes in the airspace called flight levels $\{0,1,\cdots,L-1\}$ (i.e., different elevations), a set of flights $A$ with their entry and exit coordinates $(X^{a}, Y^{a})$ and $(X'^{a}, Y'^{a})$ for flight $a\in A$ on the sector's boundary with the release time $T^a$. The goal is to assign a trajectory $P^a:t\mapsto \left(x^a(t),y^a(t)\right)$ and flight level $l^a$ to each flight subject to the given constraint $C$ on trajectory (e.g.,  limited deviation from the straight path). At the very least, $C$ includes the constraint that no two flights on the same flight level can violate the separation distance $s$ for the period $T_\text{start} \leq t \leq T_\text{end}$. Finally, note that some flights can be still en route to their exit coordinates by $T_\text{end}$. Similarly, some flights might have already passed their entry coordinates at $T_\text{start}$. 

If the CRP only includes one flight-level, we may use the term planar CRP.

\section*{\begin{flushleft}\textbf{``Cluster \& Disperse'' heuristic for conflict resolution}\end{flushleft}}
In this section, we introduce our heuristic Cluster \& Disperse for conflict resolution for aircraft in a sector. We assume there are $h$ flight levels (typically, flight levels span from 30,000ft to 44,000 ft altitude and every 1,000ft constitutes one flight level).

We make use of the following concepts in Cluster \& Disperse.
\subsection*{\textbf{1. Concepts}}
\begin{itemize}
    \item Min-Separation Distance: It is the minimum allowed distance $s$ between any two aircraft at any given time. We call a so-called loss of separation a conflict.
    \item Pos-Time: Similar to the spacetime coordinates in Physics, each Pos-Time (short for Position-Time) is a 4D coordinate for the aircraft of $(x,y,l,t)$ where $x,y$ correspond to the position on the plane, $l$ corresponds to the flight-level, and $t$ to the time.
    \item Min-Distance-Events: The set of pos-time $m^{a,b}$ where for any two flights $a,b$ on the same flight level, they are at their closest distance $d^{a,b}$ given their assigned trajectories.  Formally, if $t^{a,b}$ is the time the above happens, then $m^{a,b}$ is defined as follows.
    \begin{equation}
        d^{a,b} = \min_t \norm{                    \begin{bmatrix}
                      x^a(t) - x^b(t) \\
                      y^a(t) - y^b(t)
            \end{bmatrix}              
        }
    \end{equation}
    \begin{equation}
        t^{a,b} = \arg\min_t \norm{
           \begin{bmatrix}
                      x^a(t) - x^b(t) \\
                      y^a(t) - y^b(t)
            \end{bmatrix}                
        }
    \end{equation}
    \begin{equation}
        m^{a,b} = \{P^a(t^{a,b}), P^{b}(t^{a,b})\}
    \end{equation}
    where $P^k:t\mapsto (x^k,y^k,l^k,t)$.
    Given any set of min-distance-events, $M'$, we define $A(M')$ to be the set of flights associated with the min-distance-events in $M'$.

    \item Pos-Time Distance: Any function that defines the distance between any two Pos-Time to be later used for clustering Min-Distance-Events. We choose the function defined below
    \begin{equation}
        ptd(p,p') = 
            \begin{cases}
                \norm{
                    \begin{bmatrix}
                      x - x' \\
                      y - y' \\
                      V_0(t-t')
                    \end{bmatrix} 
                }, & \text{if } l = l'\\
                \infty & \text{otherwise},
            \end{cases}  
    \end{equation}
    for some parameter $V_0$.

    \item Clusters of conflicts: Clusters of min-distance-events that violate min-separation distance.

    \item Conflict Contribution Score: 
    There are many choices for defining a function that intuitively captures the extent to which a min-distance-event of a flight is responsible for causing a conflict. We pick a simple form where the conflict contribution score for each min-distance-event $p$ with respect to a set of min-distance-events $M'$, called $ccs(p|M')$ changes linearly with its distance to each min-distance-event $p' \in M' - \{p\}$ as follows
    \begin{equation}
        ccs(p|p') = 
            \begin{cases}
                \norm{p-p'}, & \text{if } d^{a,a'}\leq s'\\
                0 & \text{otherwise},
            \end{cases}     
    \end{equation}
    \begin{equation}
        ccs(p|M') = \sum_{p'\in M' - \{p\}} ccs(p|p'),
    \end{equation}

\end{itemize}
where $s':=s+s_0$ and $s_0$ is a margin parameter.

\subsection*{\textbf{2. Overview of Cluster \& Disperse}}
\begin{figure}[!t]
\centering
\includegraphics[trim=70 0 220 0,clip,width=0.45\textwidth]{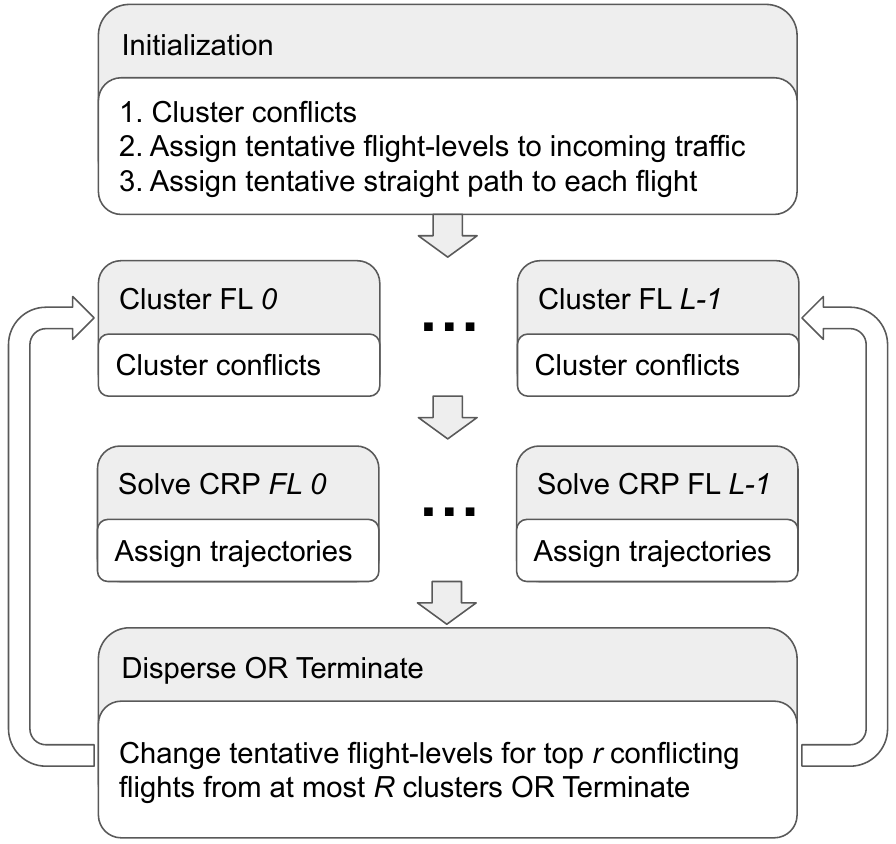}
\caption{This figure shows a visual summary of the Cluster \& Disperse heuristic.}
\label{fig_stepsAlgorithm}
\end{figure}

Cluster \& Disperse is composed of the following steps (refer to Fig. \ref{fig_stepsAlgorithm} for a visual summary).

\begin{itemize}
    \item Initialization: For the initial configuration, all flights are placed on one flight level, and assigned a direct flight trajectory between their pair of entry and exit coordinates on the sector; as if no other aircraft exists. The separation violating min-distance-events are gathered. We feed this generated data to the k-means clustering algorithm, to obtain the conflict clusters. We sort each cluster, based on an assigned conflict contribution score to each min-distance-event. For each conflict cluster, we find the associated flights with the min-distance-events. Intuitively, we would like to place flights that are in the same conflict cluster on different flight levels. Therefore, we use round robin to distribute flights between the flight levels, ignoring possible duplicate flights in the next conflict clusters (this can happen as the same flight might have produced min-distance-events in multiple conflict clusters).

    \item Clustering conflicts: In each flight-level, we cluster the separation violating min-distance-events using the k-means clustering algorithm. We store the involved flights in each cluster. This data will be later used in the Disperse step. A visual representation of this step is shown in Fig. \ref{fig_expandConflicts}.

    \item Solve planar CRPs: We solve planar CRPs for each flight level using any existing conflict resolution algorithm. If no conflict exists at any level, we skip the Disperse step and go to the Termination step. A visual representation of this step can also be seen in Fig. \ref{fig_expandConflicts}

    \item Disperse: In this step, given the trajectories from the previous step, for each flight level, we assign a conflict contribution score to each min-distance-event. We sort each cluster based on the conflict contribution scores and disperse from each cluster at most $r$ biggest contributor and in total from all clusters at most $R$ flights to the other chosen at random flight levels. Then we repeat the steps above from the clustering step for a finite number of times, $N$. A visual example of this step is shown Fig. \ref{fig_disperseFlights} and Fig. \ref{fig_clusterSorting}.
    \item Termination: Report the trajectories and flight-levels for each flight. If by any chance, any conflicts could not be resolved, report the violating flights so they can be removed from the schedule. 
\end{itemize}

In the following subsections, we will explain each step in more detail.

\subsection*{\textbf{3. Initialization}}
We initially assign the flight-level $0$ to all flights and generate straight-line trajectories 
\begin{equation}
    \begin{bmatrix}
      x^{a} \\
      y^{a} \\
    \end{bmatrix}
     = (t-T^{a})v^a + 
    \begin{bmatrix}
      X^{a} \\
      Y^{a} \\
    \end{bmatrix}
\end{equation}
\begin{equation}
    v^a = V^a\dfrac{D^a}{\norm{D^a}}
\end{equation}
\begin{equation}
    D^a = 
    \begin{bmatrix}
      X'^{a} - X^{a} \\
      Y'^{a} - Y^{a} \\
    \end{bmatrix} 
\end{equation}
where $V^a$ is the desired velocity for aircraft $a$. Based on these trajectories, the set $M_\text{init}$ of conflicting min-distance-events is 
\begin{equation}
    M_\text{init} = \bigcup_{a,b} m^{a,b}.
\end{equation}
After feeding $M_\text{init}$ to the k-means algorithm, at most $N_{init}$ clusters will be generated with $ptd$ as our Pos-Time distance function and where $N_\text{init}$ is a set parameter and $0\leq i\leq N_\text{init}-1$. For each cluster 
\begin{equation}
    M_\text{init}^i= \{p^{i_1}, p^{i_2}, \cdots, p^{i_{n(i)}}\},
\end{equation}
we sort the elements according to the $ccs$ function in descending order such that for $p^{i_k}$ and $p^{i_{k'}}$, we have $k<k'$ if and only if $ccs(p^{i_k}|M_\text{init})\geq ccs(p^{i_{k'}}|M_\text{init})$. 
Let $A_\text{init}^i$ be the associated flights with $M_\text{init}^i$ (while keeping the sorted order\footnote{For the case, where there are more than one min-distance-events in $M_\text{init}^i$ belonging to the same flight, the highest conflict contribution score will be used for determining the sort order of that flight.}) excluding the duplicate flights:
\begin{equation}
    A_\text{init}^i = A(M_\text{init}^i) - \bigcup_{j<i} A(M_\text{init}^j).
\end{equation}
\begin{equation}
    A_\text{init}^i = \{a^{i_1}, a^{i_2}, \cdots, a^{i_{n'(i)}}\},
\end{equation}

Next, using round robin, we assign flight levels according to 
\begin{equation}
l^{i_j} \gets j-1+\sum_{k<i}n'_i \mod L
\end{equation}
where $a^{i_j}\in A_\text{init}^i$.
For level $e$, $A_e$ will be the set of flights assigned to that level.

\subsection*{\textbf{4. Cluster}}
On iteration $h$, for flight-level $e$, using k-means solvers we generate  $N_e^h$ clusters $C_e^i$ of min-distance-events with $ptd$ as our distance function where $0\leq i \leq N_e^h-1$. We store the flights involved in each cluster as 
\begin{equation}
    A_e^i \gets A(C_e^i)- \bigcup_{j<i} A(C_e^j)
\end{equation}
and later use them in the Disperse step, while removing the duplicates that show in more than one cluster.

\subsection*{\textbf{5. Solve planar CRPs}}
\begin{figure}[!t]
\centering
\includegraphics[trim=0 240 373 0,clip,width=0.45\textwidth]{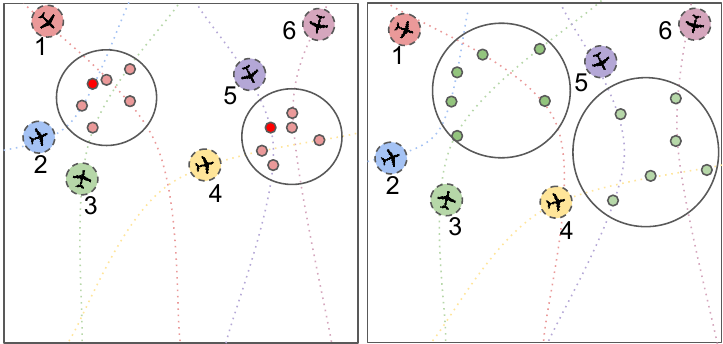}
\caption{In the left figure, the conflict clusters are detected in the Clustering step. In the right figure, new tentative trajectories are shown after a planar CRP solver is used to resolve the conflicts.}
\label{fig_expandConflicts}
\end{figure}

Using any planar CRP-solver, we solve the instance of the problem for flights $A_e$ on flight-level $e$ using any conflict resolution solver. One particular choice can be the RF-leg-based CRP-solver solver we have devised in this paper as described in the next section (Fig.  \ref{fig_expandConflicts}).

\subsection*{\textbf{6. Disperse}}
\begin{figure}[!t]
\centering
\includegraphics[trim=0 70 370 0,clip,width=0.45\textwidth]{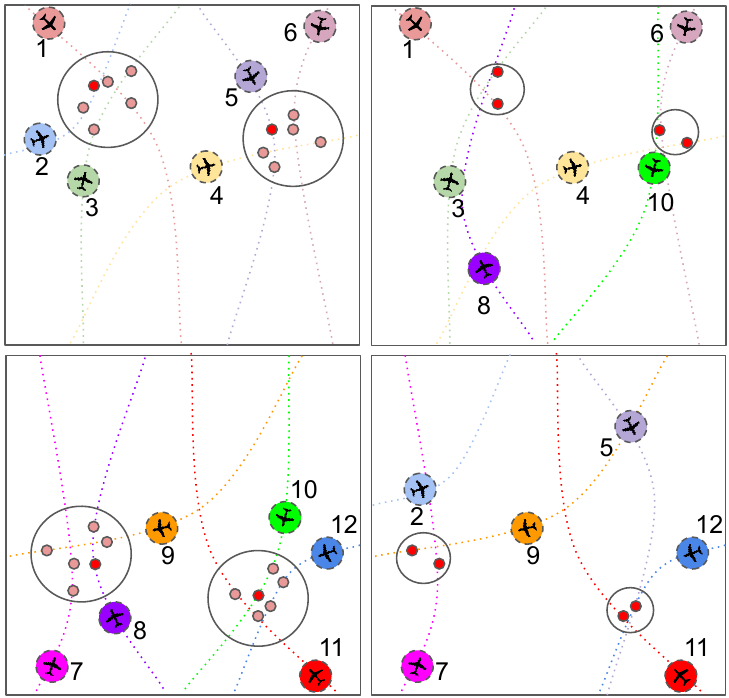}
\caption{Two horizontal planes associated with two flight-levels are shown on the left. From each flight cluster, the top contributor to the conflict is removed and added to the other flight-level. The result is shown on the right. The newly formed trajectories on these two flight-levels have fewer conflicts.}
\label{fig_disperseFlights}
\end{figure}

\begin{figure}[!t]
\centering
\includegraphics[trim=55 55 437 229,clip,width=0.45\textwidth]{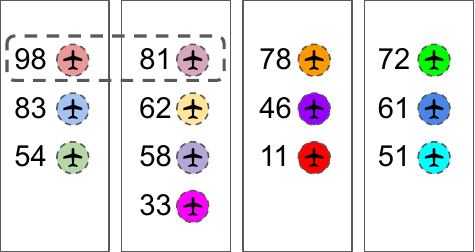}
\caption{In this figure, four clusters related to one flight-level are shown. The conflict contribution score for each flight with respect to the cluster is shown next to them. As an example, we aim to choose $R=2$ flights in total while choosing $r=1$ top conflict-inducing flight from each cluster. These chosen flights are then assigned to another flight-level.}
\label{fig_clusterSorting}
\end{figure}

Given the trajectories of the flights from the previous step for $a\in A_e$, for each flight level $e$, we obtain the new set of min-distance-events $M_e$. 
We sort all the flights involved in the conflicts in the following way. First, we assign a score $S^a$ to each flight; as a proxy for their contribution in the conflicts, as follows. 
Let $m^a$ be
\begin{equation}
    m^a= \bigcup_b m^{a,b}.
\end{equation}
Then we have
\begin{equation}
    S^a = \max_{m\in m^a} ccs(m|M_e)
\end{equation}
representing the most dangerous contribution of a flight to the conflicts among all its min-distance-events. We sort all flights based on their score and aim to select $R$ flights with higher scores from sets $A_e^i$ to be moved each to another flight level at random. In choosing these $R$ flights, we act equitably toward clusters. That is, in each sweep, each set is allocated a quota of $r$ toward forming the set of $R$ flights. Fig. \ref{fig_clusterSorting} and Fig. \ref{fig_disperseFlights} represent this step.

\section*{\begin{flushleft}\textbf{Planar CRP-solver}\end{flushleft}}
A possible choice of a planar CRP-solver is our novel algorithm RF-leg conflict resolution method. We borrow many ideas and concepts from Cluster \& Disperse in this algorithm too as will be seen shortly.

\subsection*{\textbf{1. Background on RF-leg segments}}
\begin{figure}[!t]
\centering
\includegraphics[trim=100 145 0 77,clip,width=0.45\textwidth]{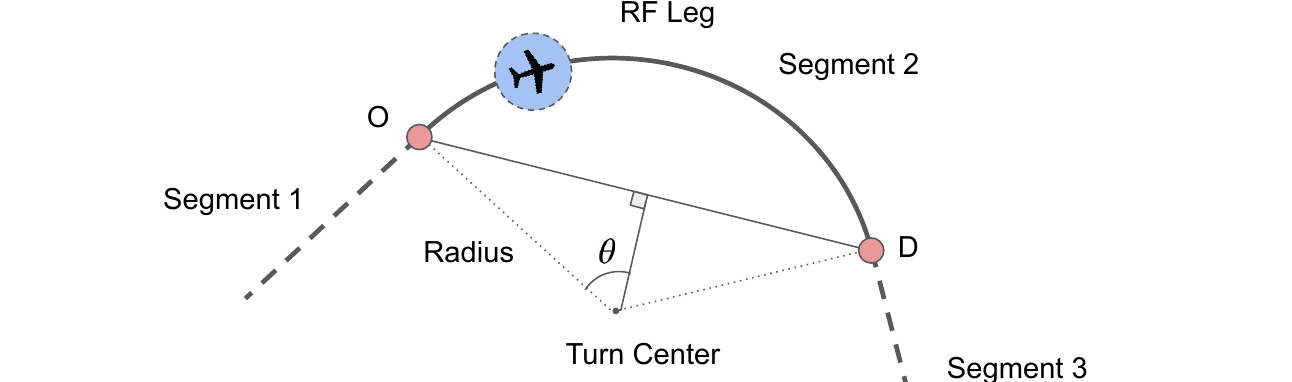}
\caption{RF-leg segments are arcs with fixed radius that connect two points. In this figure, a flight is instructed to fly over this arc to get from point O to D. The radius or the angle $\theta$ as will be used in our work gives an exact specification of the arc.}
\label{fig_RF_Radius}
\end{figure}

Radius-to-Fix legs have recently become a routine part of trajectories in aviation where a flight travels along an arc to get from point A to B, instead of the direct path \cite{miller2011integration}. Providing the center of the arc or the angle as shown in Fig. \ref{fig_RF_Radius} can uniquely specify the trajectory. 

\subsection*{\textbf{2. Overview of the RF-leg planar CRP algorithm}}
We give a brief overview of RF-leg planar CRP algorithm here and explain each step in more depth in the next few subsections. In RF-leg CRP algorithm, each aircraft is assigned a constant speed throughout its flight. Initially, a straight flight trajectory is assigned to each flight. Our goal is to assign the angle $\theta^a$ for each flight $a\in A$ such that $\theta_\text{low}\leq \theta^a\leq \theta_\text{high}$ and the resulting trajectories are conflict-free. We take the following steps accordingly. First, we find the conflict clusters of min-distance-events $C_j$'s using the k-means algorithm. Then we assign an intra-cluster conflict contribution score to each min-distance-event in a cluster. For each cluster $C_j$, we run the gradient-descent algorithm to find $\theta^a$ for each flight $a$ that has a corresponding min-distance-event in $C_j$. We run the gradient descent until either we get close to local minima or resolve all the conflicts.

\subsection*{\textbf{3. Clustering step}}
We first extract the min-distance violating min-distance-events and use the k-mean algorithm to find a total of $N_\text{RF}$ clusters in $M$, the set of illegal min-distance-events, defined as 
\begin{equation}
M = \bigcup_{a,b} m^{a,b}
\end{equation}
and where $N_\text{RF}$ is a given parameter.

\subsection*{\textbf{4. Gradient descent}}
In this step, we aim to minimize the total intra-cluster conflict contribution score for each cluster of min-distance-events. We choose to use the same $ccs$ function from the previous section. More formally, we seek local minima for the total conflict score ($tcs$) function of the set of min-distance-events $M'$
\begin{equation}
tcs(M'):=\sum_{p}ccs(p|M').
\end{equation}
We do this by finding a locally optimal $\theta^a$ for $a\in A(M')$ using the stochastic gradient descent approach with an adaptive learning rate. Initially, we assign all flights in $A(M')$ to be the straight path trajectories; that is we set initial $\theta^a=0$. Next, we update $\theta^a$ according to the following recursive assignment rule:
\begin{equation}
    \theta^{A(M')} \gets \theta^{A(M')} - \eta\cdot\dfrac{\nabla tcs}{\norm{\nabla tcs}}
\end{equation}
where $\eta$, known as the learning rate is a positive parameter that adapts dynamically depending on the size of the gradient, $\nabla tcs(M')(\theta^{A(M')})$ (i.e. the rate of improvement to the cost function). 

It is worth mentioning the following aspects of our gradient descent algorithm.
\begin{itemize}
    \item Termination: In each iteration, if the improvement in the value of $tcs$ is less than some set threshold parameter $T_{GD}$, that is $0 \leq \Delta tcs \leq T_{GD}$, then we will terminate the execution of the algorithm.
    \item Adaptive learning rate: In each iteration, if the improvement in the value of $tcs$ is not fast enough according to some set threshold parameter $T'_{GD}$ (larger than $T_{GD}$), that is $0 < \Delta tcs \leq T'_{GD}$, then we will increase the learning rate $\eta\gets \eta W_U$ where $1<W_U$ is a parameter.
    \item Backtracking: In each iteration, if the value of $tcs$ worsens, we backtrack to the previous solution and decrease the weight $\eta\gets \eta W_D$ where $W_D<1$ is a parameter and try again.
    \item Out of bound: If any component of $\theta^{A(M')}$, that is $\theta^{a'}$ for $a'\in A(M')$ gets out of the range of $[\theta_\text{low},\theta_\text{high}]$ in one of the iterations, we will set it back to the closest value in the range (thinking linearly about the angular range rather than periodic).
\end{itemize}

\section*{\begin{flushleft}\textbf{Case study}\end{flushleft}}

\subsection*{\textbf{1. Application to a sector}}
We consider an instance of CRP with 12 flight levels and a rectangular sector with dimensions 54nmi by 64.8nmi. All aircraft fly at a speed of 533kn. On every edge, we allow entries and exits at designated points placed uniformly every 5.4nmi apart. We release 320 flights over 1 hour ($T_\text{start}=0 \text{hr}$ and $T_\text{end}=1\text{hr}$) from the boundaries of the sector at designated entry points. The entry and exit points for each flight are generated uniformly at random with the stipulation that they cannot be located on the same edge of the rectangle-shaped boundary. All flight release times are a multiple of 0.02 hours and at most one flight can be released from a designated point at any point in time. For calculating the min-distance-events, we discretize the time with the time step $dt=2.5s$.

We set the parameters of Cluster \& Disperse, as well as RF-leg planar CRP solver as follows.
\begin{itemize}
    \item $s_0 = 0.625\text{nmi}$, $s = 5\text{nmi}$
    \item $V_0 = 533kn$, $V^a = 533kn$
    \item $T_{GD} = 10^{-7}$, $T'_{GD} = 10^{-3}$
    \item $W_U = 1.5$, $W_D = 0.5$
    \item $N = 10$
    \item The parameters for the number of clusters are set as follows \begin{equation}
        N_\text{init}=\left\lfloor\dfrac{\#\text{min-distance-events}}{5}\right\rfloor.
    \end{equation}
    Also, $N_e^h$ and $N_{RF}$ are similarly set in each iteration.
    \item $\theta_\text{low}=-25\degree,\theta_\text{high}=+25\degree$
\end{itemize}

\subsection*{\textbf{2. Numerical results and algorithm performance}}

\begin{figure}[!t]
\centering
\includegraphics[trim=70 70 70 60,clip,width=0.45\textwidth]{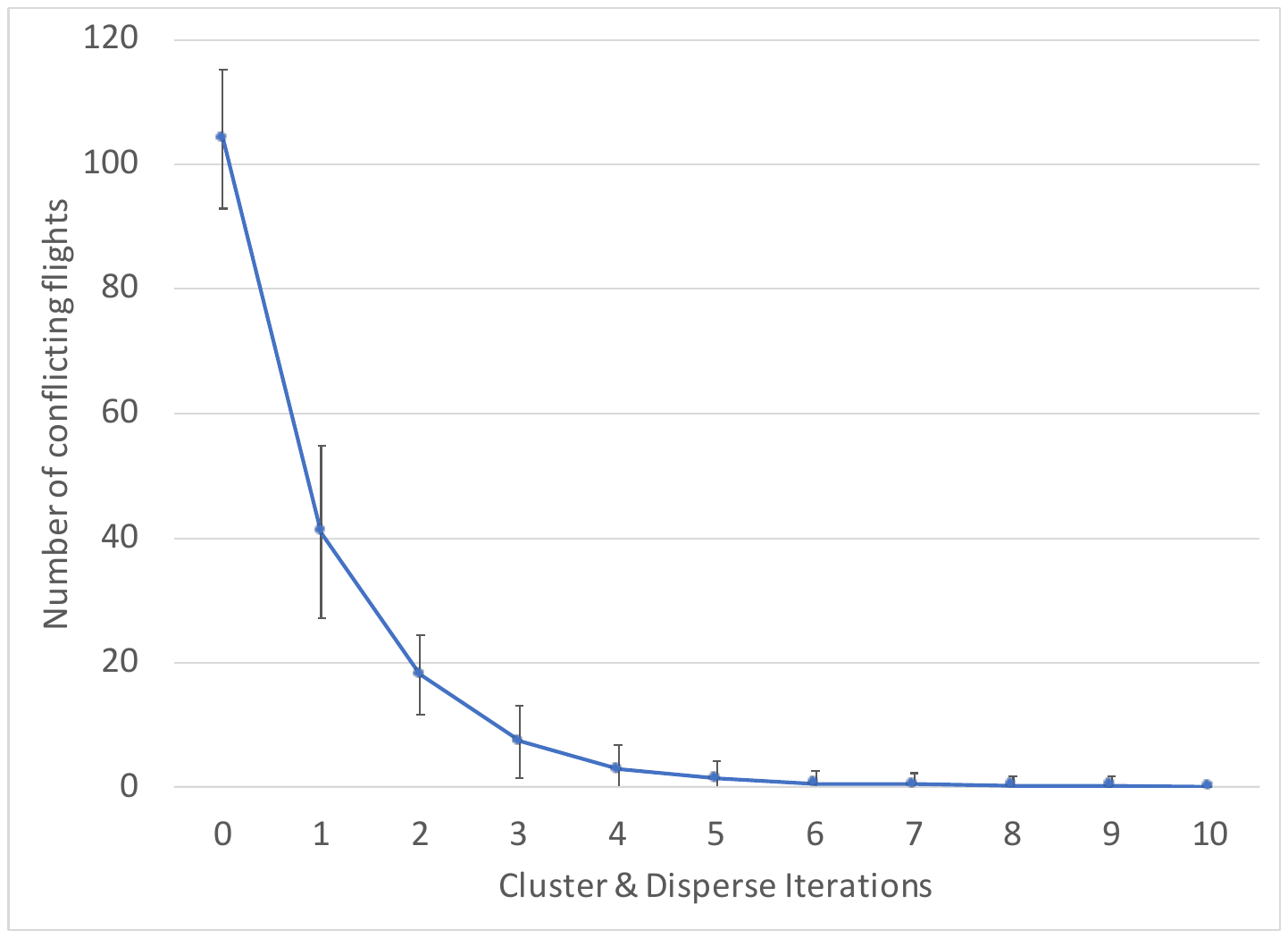}
\caption{In this figure, the average number (over all 20 instances) of the conflicting flights per the number of iterations is shown. A partially improved solution will be available after each iteration until 
 all the conflicts are finally resolved. On average at the beginning about $1/3$ of the flights have at least one conflict with other flights.}
\label{fig_countConflictFlights}
\end{figure}

We run the Cluster \& Disperse together with RF-leg planar CRP solver. In Fig. \ref{fig_countConflictFlights}, we see how the average number of conflicting flights decreases with each iteration of Cluster \& Disperse. On average at the beginning about $1/3$ of the flights have at least one conflict.

\begin{figure}[!t]
\centering
\includegraphics[trim=70 70 70 60,clip,width=0.45\textwidth]{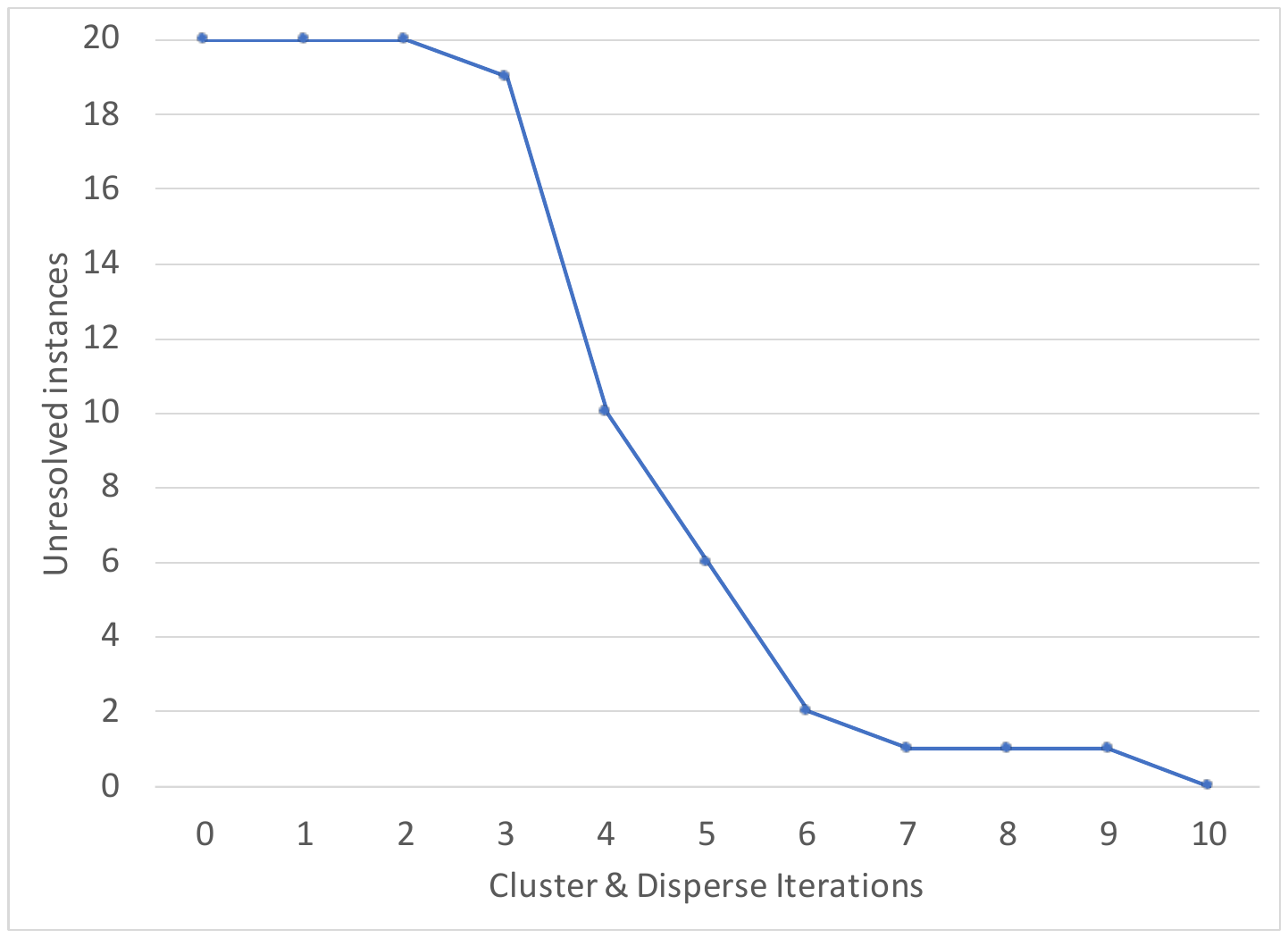}
\caption{In this figure, the number of instances of CRP that remain unresolved per the number of iterations of Cluster \& Disperse is shown.}
\label{fig_unresolvedInstances}
\end{figure}

Applying Cluster \& Disperse together with RF-leg CRP algorithm, all conflicts will be resolved among the 20 scenarios we considered by at most the 10th iteration (Fig. \ref{fig_countConflictFlights}). However, most scenarios are solved earlier than that as can be seen in Fig. \ref{fig_unresolvedInstances}. In fact, 70\% of the scenarios are resolved by the 5th iteration.

\begin{figure}[!t]
\centering
\includegraphics[trim=60 80 70 60,clip,width=0.45\textwidth]{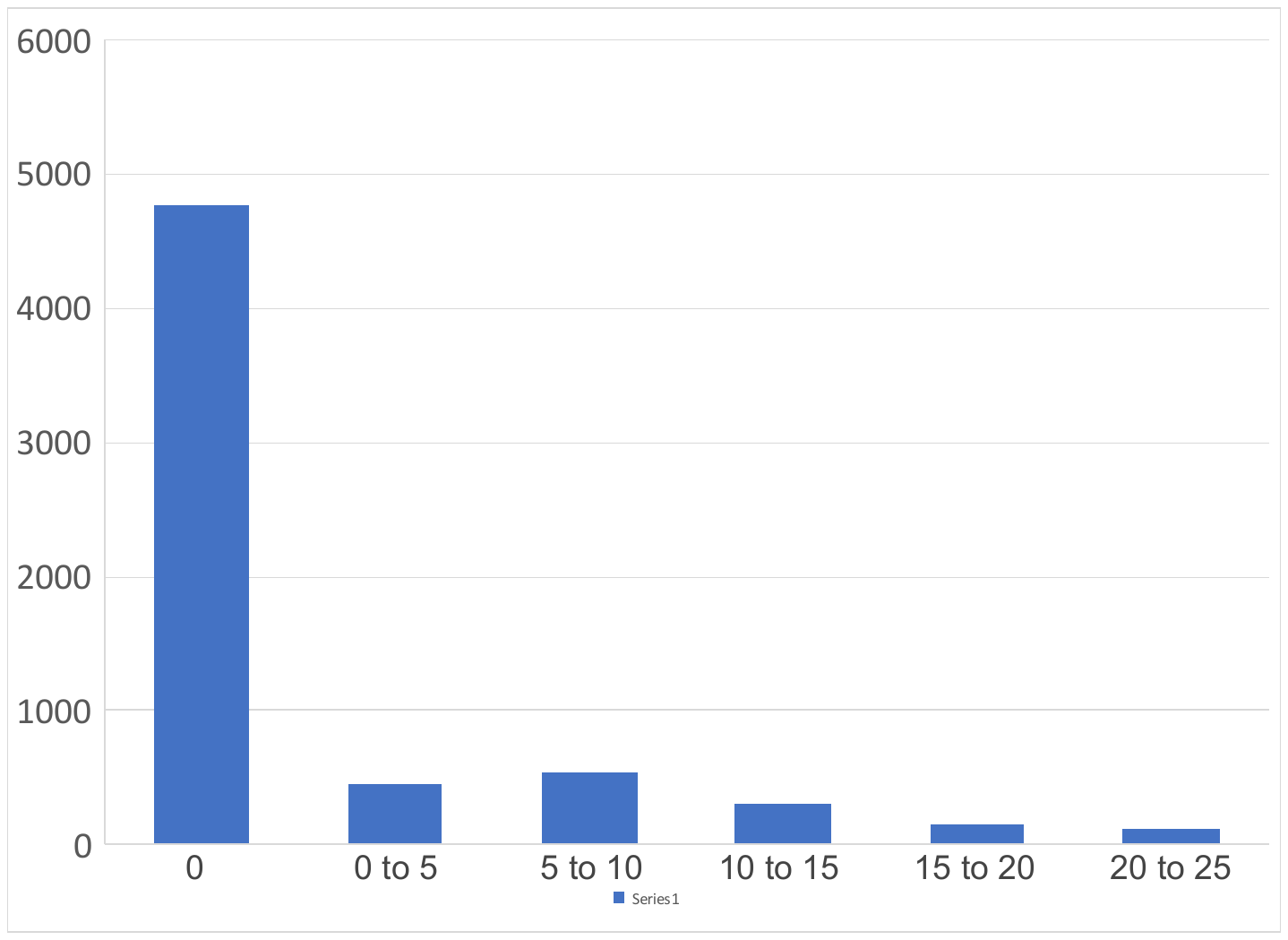}
\caption{In this histogram, the angle for the flights are shown. About 3/4 of the flights will fly on a straight path while the remaining flights may need at most a 25-degree deviation to avoid conflict. To clarify, this is not the heading, but half the path's arc angle.} 
\label{fig_anglesHistogram}
\end{figure}

\begin{figure}[!t]
\centering
\includegraphics[trim=60 80 70 60,clip,width=0.45\textwidth]{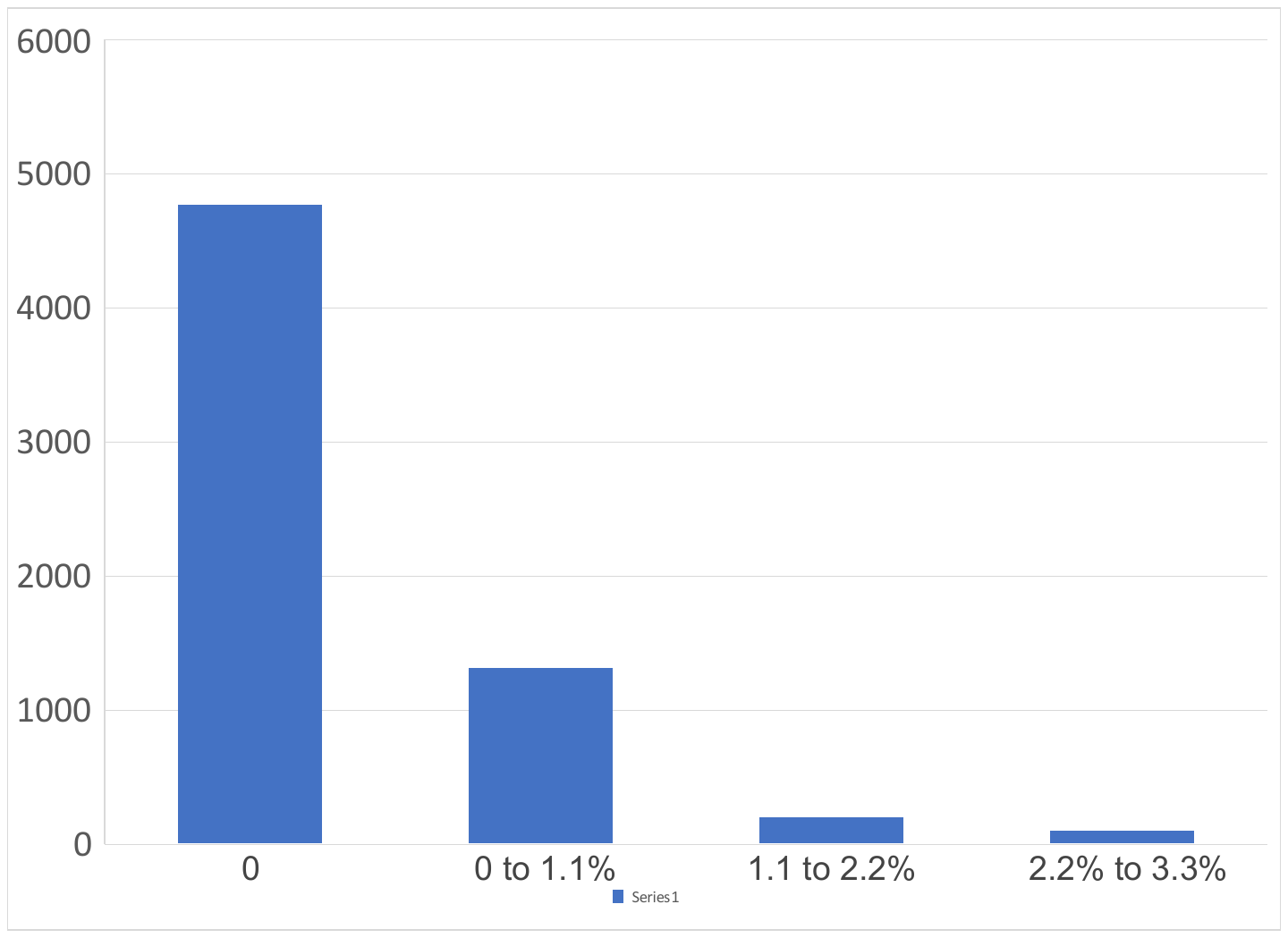}
\caption{In this histogram, the path for each flight is lengthened by 0 to at most 3.3\% depending on the severity of the maneuver. As can be seen, almost 75\% of the flights will fly on a straight path while almost all the remaining flights need at most 1.1\% of extension to their path to avoid conflict. Overall, the path for the flights compared to the straight path will be lengthened only by a negligible average factor of $1.0015$.}
\label{fig_extraDistHistogram}
\end{figure}

In terms of the efficiency of our solutions, using Cluster \& Disperse with RF-leg planar CRP solver, 3/4 of flights will fly on a straight path. Of the remaining 1/4, most are required to deviate from their path by an RF-leg angle of at most 5 or 10 degrees. The histogram in Fig. \ref{fig_anglesHistogram} shows the distribution of the angles required for successful conflict resolution. The extra distance that each flight needs to go in terms of a percentage point is shown in Fig. \ref{fig_extraDistHistogram}. On average, the path for the flights will be lengthened by less than $1/6$ of 1\%, that is $0.0015$ yielding an efficiency $99.85\%$ (which we define as the inverse of the previous ratio).

The average peak number of simultaneous flights among the 20 instances is $45\pm 3$ and it ranged between $40$ and $51$ among the samples. This translates into an average peak density of 154 aircraft per 10,000sq nmi. 

To put this in context, according to FAA, the average peak number of flights in the US is 5300 which yields an upper bound of 18 aircraft per 10,000sq nmi across the US airspace at any given minute. The distribution variations between sectors however can be large \cite{FAA22}. In particular, using the live statistics on the website www.ztlartcc.org, we monitored once per hour the number of aircraft present in Atlanta (ZTL) Air Route Traffic Control Center (ARTCC) starting on Monday, February 13, 2023 at 5pm PST for 48 hours. This ARTCC is considered the busiest ARTCC in the United States. From our observations, the peak number of aircraft present were 22. Even if we make the unrealistic assumption that all these aircraft were concentrated in the smallest sector among any of the possible tens of sectors within this ARTCC, our calculations yields an upperbound of 62 aircraft per 10,000sq nmi. This is still far lower than the peak density that we have considered in our work, as mentioned above.

We ran our code on a MacBook Pro with the Apple M1 Max chip and 32GB of memory. Our algorithms are implemented in Python (single-threaded, hence utilizing one CPU core) and the run time for our algorithms is shown in table \ref{tab:table1}. Except for the small initialization, the share of the Cluster \& Disperse in the run-time is negligible. The serial average run-time for our algorithm is $400\pm 143s$. 
Each iteration of the overall algorithms combined takes $74\pm21s$ and each instance takes $5\pm1.6$ iterations to resolve. In each iteration, we churn out a partial solution with substantially fewer conflicts. If the time is tight, the flight controller can rely on one of these early solutions while manually separating likely at most one flight -if any- at each level at any moment (Note that from \ref{fig_countConflictFlights}, after the 1st iteration, only 40 conflicting flights exist on average over 12 flight-levels over 1 hour).

While we did not implement a parallel version of our algorithm, the simplest implementation will be to solve for each flight level separately on a separate CPU core for maximum speed up. In addition to the approximately $10s$ initialization step, we give likely pessimistic estimates of the parallel average run-time of $7.9s$ per iteration and about $40s$ for the entire iterations. 

\begin{table}[!t]
\caption{Run-time performance of Cluster \& Disperse combined with RF-leg planar CRP solver\label{tab:table1}}
\centering
\begin{tabular}{|c||c|}
\hline
Item & Run-time (s)\\
\hline
Initialization & $9.8\pm0.1$\\
\hline
Serial average run-time & $400\pm 143$\\
\hline
Estimated parallel average run-time & $40$\\
\hline
\end{tabular}
\end{table}

\subsection*{\textbf{3. Performance of the existing algorithms}}
The results presented above compare favorably to prior work. While the conditions for each study are different, thus making a quantitative comparison to our algorithms difficult, we want to provide some qualitative perspective on our algorithm's performance relative to prior work. 


In \cite{krozel2001system}, three methods are devised for the conflict resolution problem. The highest density considered is 25 aircraft per 10,000sq nmi in a circular sector of radius 100nmi.
Over the span of 50 minutes, 80 simultaneously flying aircraft are simulated. The efficiencies achieved for this density (measured as the extra travel distance added as a proxy for fuel cost) are 0.97, 0.95, and 0.93 for the three methods, respectively.

In \cite{vela2009mixed}, an irregularly shaped sector with 3 flight levels is considered where the sector roughly occupies half of a rectangle of size 130nmi by 170nmi. The peak simultaneous aircraft count is 44 when 300 aircraft per hour enter the sector, constituting an approximate peak density of 20 aircraft per 10,000sq nmi. The distribution of the flights is not uniform, and the flights considered all travel westbound. The algorithm runs ILOG CPLEX version 10.0 on a system that uses four parallel 2GHz CPU with a memory of 3GB and the computation time is limited to 600s. As a result of maneuvers, the cost is increased on average by 2\% for each aircraft and the number of infeasible flights on average is 0.3.

In \cite{lehouillier2017solving}, a square sector of 50nmi per side is considered. In one of the scenarios, 35 aircraft fly simultaneously giving a density of 140 aircraft per 10,000 sq nmi. It takes about 130s to solve the MILP model which incorporates various uncertainties in the trajectories. The algorithm is implemented in the programming language C++ and uses CPLEX 12.5.1.0 as a library to solve the model. The hardware setup is a 3.4 GHz Intel i7-3770 CPU with 8 GB of memory. 

In \cite{wang2020cooperation}, a circular sector of 100nmi radius is considered with 5 flight levels. In this sector, 60 simultaneously flying aircraft are considered, yielding a density of 19 aircraft per 10,000 sq nmi. The conflict resolution algorithm solves 10 instances at this density and it takes about 140s to find a solution. The algorithm was run on Debian GNI/Linux 9.4 with a 3.4 GHz Intel Xoen 8-core CPU and 16 GB of memory. A 300s computation time is imposed on the algorithm.

We would like to note that in addition to the density of the flights, the average length of flights is also an important consideration since a longer flight will provide more opportunities for conflict.

\section*{\begin{flushleft}\textbf{Conclusion}\end{flushleft}}
In this work, we identified clusters of minimum distance events (i.e., 4D position and time of a conflict) in each flight level. Our goal was to explore the neighborhood of trajectories and flight-levels by shuffling problematic flights between various flight-levels. To pinpoint the problematic flights, we looked at the conflict clusters in each flight-level and moved a few of the flights with the highest social force scores within their cluster to another flight-level. More precisely, in our case study, we moved the top conflict flights from two clusters in each step. Therefore, we removed at most two flights from each flight-level and added them to another flight-level in each iteration. This was at the core of the Cluster \& Disperse heuristic. Within that heuristic, we used the subroutine RF-leg planar CRP solver to solve the conflict resolution problem independently for each flight-level. Here again, we used a similar idea by clustering the minimum distance events and expanding them using a social force. To do so, we used the gradient descent algorithm by adjusting the angles of the curvatures in the produced trajectory arcs. We run these two algorithms in combination for 20 different instances and in every instance all the conflicts were resolved within at most 10 iterations of Cluster \& Disperse. We achieved an efficiency of 99.85\% in terms of the deviation of the trajectories from the straight path. In 95\% of the cases, our serial run-time was less than 10 minutes. It is a possibility to achieve an order of magnitude speedup from parallelization. In each iteration, our algorithms created a partial solution with a substantially reduced number of conflicts in case a quick solution was needed by a traffic controller. Finally, we cited some of the performance results from some notable papers from the literature to provide a perspective on our numerical results. 

\section*{\begin{flushleft}\textbf{Future work and discussion}\end{flushleft}}
There is great potential in combining Cluster \& Disperse with existing planar CRP solvers in the literature and improve their conflict resolution performance. 

The run time of our algorithms can be improved by about an order of magnitude if planar CRP solvers (in our case, RF-leg planar CRP solver) are run in parallel for different flight-levels. The use of a multi-threaded programming language or library can make this task easier. In addition, in each step of the RF-leg planar CRP algorithm, each cluster is optimized sequentially. It is plausible that by running these optimizations in parallel, another significant speed-up factor can be achieved. However, since two clusters can share the same flight, some tie-breaking rules will be needed.

Another potential speed-up for the algorithm may come from keeping the solution obtained from the previous step for each flight level and starting from there in each iteration. It is conceivable that adding at most two other flights to each flight-level may not change the obtained solution drastically and hence by keeping the previous solution and using it as a starting point, a vast amount of calculation in the gradient descent algorithm can be salvaged.

In our implementation, we enumerate all the pairwise distances between aircraft to detect loss of separation. This is a time-consuming part of our CRP-2D solver algorithm. Also adding noise and uncertainty to the model will enrich the model while making these calculations much more difficult as the aircraft position will be an envelope and not a point anymore. Fortunately, the loss of separation can be detected in more efficient ways than the brute force approach (i.e., examining all the pairwise distances) using a variety of well-known methods such as Graham's algorithm and GJK algorithm and the interested readers are referred to  \cite{graham1972efficient, bergen1999fast, gilbert1988fast}.

We also experimented with a higher number of aircraft up to 640 flights entering the sector, with an average peak number of $82\pm5$ simultaneous flights. The serial run-time deteriorated to an average of 4 hours. We find it interesting to report that in 50\% of the 20 instances we considered, all conflicts were resolved. In another 35\%, removal or manual separation by flight controller of only 1 aircraft would suffice to resolve the conflict. In the remaining 3 cases comprising 15\% of the total, the removal of at most  5, 7, and 18 aircraft would resolve the conflicts.

In choosing a reasonable number of clusters, we looked at the number of min-distance-events as a clue for picking a reasonable number. A slower but more precise approach can be to use the elbow method in determining the number of clusters, as conceived in \cite{thorndike1953belongs}. An alternative approach can be to use the ANOVA F-test statistic \cite{hahs2013introduction}.

Our framework is very malleable for adding other constraints. A useful constraint to explore would be to limit the range of change in flight-levels for heavier aircraft (by looking at their flight level in the previous sector). It will be interesting to use realistic data and study the performance of the algorithm with this new set of constraints. Furthermore, we do not consider vertical maneuvers and are assigning flight-levels from the point of entry. In other words, the flight in the previous sector must adjust flight-level before entering our sector. It will be an interesting direction to either integrate vertical maneuvers inside our sector or to extend this work to multi-sector in a way that minimizes the change of flight-level along the route.

\section*{\begin{flushleft}\textbf{Acknowledgments}\end{flushleft}}
We would like to thank Zhenyu Gao for his helpful feedback on this research project.


 
\bibliography{IEEEabrv,refs}
%

\bibliographystyle{IEEEtran}



\newpage

\vfill

\end{document}